%% file: _main.tex
\input{_constants}
\cameraready

\pdfoutput=1
\documentclass[10pt,twocolumn,letterpaper]{article}
\input{cvpr_header}

\unless\ifarxiv \myexternaldocument{_supplementary} \fi

\begin{document}
\title{\paperTitle}
\author{\authorBlock}
\maketitle

\def\tabA{
\begin{table*}[t]
\label{tabA}
\centering
\caption{Robustness values measured for various data augmentation image datasets~\cite{PIXMIX}. The experiments have been conducted on CIFAR-10-C and -100-C by using MoireDB within the framework of PixMix. Lower is better for the listed scores.}
\begin{tabular}{ll|ccccccc}
\toprule[0.8pt]
Dataset&  & Baseline & Fractal arts & FVis & FractalDB & VisualAtom & MoireDB \\ \hline
& Corruptions & 26.4 & 10.8 & \underline{9.5} & 11.9 & 10.8 & \textbf{9.4} \\
\multirow{-2}{*}{CIFAR-10-C} & Adversaries & 91.3 & \underline{82.0} & \textbf{78.6} & 92.2 & 93.9 & 84.2 \\ \hline
& Corruptions & 50.0 & 33.3 & \textbf{30.3} & 35.0 & 33.4 & \underline{30.9} \\
\multirow{-2}{*}{CIFAR-100-C} & Adversaries & 96.8 & \underline{93.2} & \textbf{92.3} & 98.5 & 98.5 & 95.4 \\
\bottomrule[0.8pt]
\label{tabA}
\end{tabular}
\end{table*}
}

\def\tabNoise{
\begin{table}[t]
    \label{tabNoise}
    \centering
    \caption{Image classification accuracy for CIFAR-100-C images corrupted by various types of noise. Lower is better for the listed scores.}
    \begin{tabular}{l|ccc}
        \toprule[0.8pt]
                   & Gaussian & Shot & Impulse \\ \hline
        Fractal arts   & 43.4     & 36.7 & \underline{30.4}    \\
        FVis       & \textbf{31.2}     & \textbf{28.6} & \textbf{29.8}    \\
        FractalDB  & 52.0     & 43.9 & 31.6    \\
        VisualAtom & 45.7     & 39.5 & 32.6    \\
        MoireDB    & \underline{38.6}     & \underline{34.6} & 34.1    \\ 
        \bottomrule[0.8pt]
    \end{tabular}
    \label{tabNoise}
\end{table}
}

\def\tabBlur{
\begin{table}[]
    \label{tabBlur}
    \centering
    \caption{Image classification accuracy for CIFAR-100-C images corrupted by various types of blurring. Lower is better for the listed scores.}
    \begin{tabular}{l|cccc}
        \toprule[0.8pt]
                   & Defocus & Glass & Motion & Zoom \\ \hline
        Fractal arts   & 26.3    & 60.4  & 32.4   & 28.7 \\
        FVis       & 26.4    & \underline{45.8}  & \textbf{31.0}   & 30.0 \\
        FractalDB  & \textbf{25.3}    & 64.1  & \underline{31.3}   & \textbf{27.4} \\
        VisualAtom & 26.5    & 48.4  & 32.5   & 28.8 \\
        MoireDB    & \underline{25.7}    & \textbf{39.8}  & \textbf{31.0}   & \underline{28.1} \\ \bottomrule[0.8pt]
    \end{tabular}
    \label{tabBlur}
\end{table}
}

\def\tabWea{
\begin{table}[]
    \label{tabWea}
    \centering
    \caption{Image classification accuracy for CIFAR-100-C images corrupted by various types of weathering. Lower is better for the listed scores.}
    \begin{tabular}{l|cccc}
        \toprule[0.8pt]
                   & Snow & Frost & Fog  & Brightness \\ \hline
        Fractal arts   & 30.0 & 31.6  & 26.9 & \underline{22.1}       \\
        FVis       & 29.2 & \underline{27.8}  & 27.1 & 22.8       \\
        FractalDB  & 28.4 & 32.4  & \textbf{22.1} & \textbf{21.5}       \\
        VisualAtom & \underline{28.0} & 28.2  & \underline{26.1} & 22.8       \\
        MoireDB    & \textbf{27.3} & \textbf{26.9}  & 28.2 & 23.4       \\ \bottomrule[0.8pt]
    \end{tabular}
    \label{tabWea}
\end{table}
}

\def\tabDig{
\begin{table}[]
    \label{tabDig}
    \centering
    \caption{Image classification accuracy for CIFAR-100-C images corrupted by various types of digital transformations. Lower is better for the listed scores.}
    \begin{tabular}{l|cccc}
        \toprule[0.8pt]
                   & Contrast & Elastic & Pixelate & JPEG \\ \hline
            Fractal arts   & \textbf{24.0}     & 31.6    & 37.7     & \underline{36.7} \\
        FVis       & \underline{24.3}     & 32.1    & \underline{33.7}     & \textbf{34.4} \\
        FractalDB  & 24.7     & 31.1    & 43.5     & 46.0 \\
        VisualAtom & 26.0     & \underline{31.0}    & 38.4     & 38.4 \\
        MoireDB    & 26.2     & \textbf{30.2}    & \textbf{28.0}     & 41.5 \\ \bottomrule[0.8pt]
        \end{tabular}
    \label{tabDig}
\end{table}
}

\input{00_abstract}
\input{01_intro}
\input{02_related}
\input{03_method}
\input{04_experimental}
\input{05_conclusions}

{\small
\bibliographystyle{ieeenat_fullname}
\bibliography{11_references}
}

\ifarxiv \clearpage \appendix \input{12_appendix} \fi

\end{document}

%% file: _constants.tex
\def\paperID{0000}
\def\confName{CVPR}
\def\confYear{2025}

\def\paperTitle{MoireDB: Formula-generated Interference-fringe Image Dataset}

\def\authorBlock{
    Yuto MATSUO\dag,\ddag \ \ Ryo HAYAMIZU\dag,\ddag \ \ Hirokatsu KATAOKA\ddag,$\diamondsuit$ \ Akio NAKAMURA\dag \\
    \dag Tokyo Denki University \\ \ddag National Institute of Advanced Industrial Science and Technology (AIST) \\ $\diamondsuit$ University of Oxford \\
    {\tt\small \ matsuo.y@is.fr.dendai.ac.jp}
}

\newif\ifreview 
\newif\ifarxiv 
\newif\ifcamera \newcommand{\cameraready}{\cameratrue}
\newif\ifrebuttal 

%% file: cvpr_header.tex
\ifreview \usepackage[review]{cvpr} \fi
\ifarxiv \usepackage[pagenumbers]{cvpr} \fi
\ifrebuttal \usepackage[rebuttal]{cvpr} \fi
\ifcamera \usepackage{cvpr} \fi

\input{_macros}  

\usepackage{xr-hyper}

\makeatletter
\newcommand*{\addFileDependency}[1]{
  \typeout{(#1)}
  \@addtofilelist{#1}
  \IfFileExists{#1}{}{\typeout{No file #1.}}
}

\makeatother
\newcommand*{\myexternaldocument}[1]{
    \externaldocument{#1}
    \addFileDependency{#1.tex}
    \addFileDependency{#1.aux}
}

\definecolor{cvprblue}{rgb}{0.21,0.49,0.74}
\usepackage[pagebackref,breaklinks,colorlinks,allcolors=cvprblue]{hyperref}
\usepackage[capitalize]{cleveref}
\crefname{section}{Sec.}{Secs.}
\crefname{table}{Table}{Tables}
\crefname{figure}{Fig.}{Figs.}

\ifarxiv \crefname{appendix}{App.}{Apps.}
\else \crefname{appendix}{Suppl.}{Suppls.} \fi

%% file: _macros.tex
\usepackage{graphicx}	
\usepackage{amsmath}	
\usepackage{amssymb}	
\usepackage{booktabs}
\usepackage{times}
\usepackage{microtype}
\usepackage{epsfig}
\usepackage{caption}
\usepackage{float}
\usepackage{placeins}
\usepackage{color, colortbl}
\usepackage{stfloats}
\usepackage{enumitem}
\usepackage{tabularx}
\usepackage{xstring}
\usepackage{multirow}
\usepackage{xspace}
\usepackage{url}
\usepackage{subcaption}
\usepackage{xcolor}
\usepackage[hang,flushmargin]{footmisc}

\ifcamera \usepackage[accsupp]{axessibility} \fi

% ...

\ifarxiv  \fi

\newcommand{\R}[1]{{%
    \textbf{%
        \ifstrequal{#1}{1}{\textcolor{red}{R#1}}{%
        \ifstrequal{#1}{2}{\textcolor{blue}{R#1}}{%
        \ifstrequal{#1}{3}{\textcolor{magenta}{R#1}}{%
        \ifstrequal{#1}{4}{\textcolor{teal}{R#1}}{%
                           \textcolor{cyan}{R#1}%
        }}}}%
    }%
}}

%% file: 00_abstract.tex
\begin{abstract}

Image recognition models have struggled to treat recognition robustness to real-world degradations. In this context, data augmentation methods like PixMix improve robustness but rely on generative arts and feature visualizations (FVis), which have copyright, drawing cost, and scalability issues. We propose MoireDB, a formula-generated interference-fringe image dataset for image augmentation enhancing robustness. MoireDB eliminates copyright concerns, reduces dataset assembly costs, and enhances robustness by leveraging illusory patterns. Experiments show that MoireDB augmented images outperforms traditional Fractal arts and FVis-based augmentations, making it a scalable and effective solution for improving model robustness against real-world degradations.

\end{abstract}

%% file: 01_intro.tex
\section{Introduction}
\label{sec:intro}


Image recognition techniques, particularly those based on deep learning, are promising for real-world applications; however, image classification using deep learning models is said that less robust to diverse real-world degradations than that of human visual perception~\cite{ImageNet-C,ImRobust}.
Consequently, as the accuracy of image recognition models improves, the goal of increasing classification robustness with respect to real-world degradations is widely seen as one of the central challenges in image recognition with deep learning.

One promising technique for improving the robustness of image recognition models for classification task is \textit{data augmentation} such as Mixup~\cite{Mixup} and CutMix~\cite{CutMix}.
Using these data augmentation methods, we can increase image counts while reducing overfitting, thus potentially improving robustness.
The proposed data augmentation method known as PixMix~\cite{PIXMIX} extends training datasets with images taken from mixing sets by combining real and synthetic images. The mixing is done both additively or multiplicatively. PixMix pipeline achieves improvements in both robustness and image classification accuracy.

\input{figs/en-fig1}

The mixing set images used by PixMix include mathematically generated Fractal arts and feature visualizations (FVis).
Fractal arts are collected on DeviantArt. The images are visually diverse.
FVis are collected using OpenAI Microscope. The images are created from convolutional neural networks (CNNs) such as AlexNet~\cite{alexnet}, VGGNet~\cite{vggnet}, and ResNet~\cite{resnet}, which are basically pre-trained on ImageNet~\cite{distill,FVis}.

However, the use of these images entail at least three practical disadvantages:
i) Some of the human-designed digital patterns and generative arts
are protected by copyright, and thus commercial use of PixMix
data augmentation remains questionable.
ii) Generating FVis requires multiple CNNs trained on large image
datasets, and is thus a high-cost operation for assembling images into datasets.
iii) For both Fractal arts and FVis the number of
images that may be feasibly assembled into a mixing set is limited in practice.

All of these problems may be eliminated by auto-generating data augmentation images from mathematical formulas.
For example, in the methods named ``Formula-driven Supervised Learning (FDSL)'', we can construct large-scale image datasets which are privacy-safe, copyright-free, and light collecting images costs. 
In FDSL context, Fractal DataBase (FractalDB), Shader, etc. are proposed.
As discussed in the original PixMix paper,
tests to characterize the performance of
FractalDB and Shader, indicate that formula-generated images are less effective
for improving robustness than Fractal arts and FVis.

Thus, the aim of the present study is to devise a strategy for
constructing a formula-generated mixing set that can improve the robustness of the classification model compared to fractal / FVis.
Our data augmentation procedure is the same as that of PixMix,
but here we propose a novel family of data augmentation
images that promise improved robustness:
\textit{interference-fringe images}, for which 
we use the term \textit{MoireDB} images and their constructed \textit{Moir\'e DataBase (MoireDB)}.

The idea of generating Moir\'e images
is motivated by the hypothesis that using
illusory images for data augmentation should
tend to increase robustness; this hypothesis, in
turn, is based on the close relation between Moir\'e images
and optical illusions, as well as on the known fact that
deep learning models trained on such images exhibit
increased robustness.
We construct MoireDB, a repository
of automatically formula-generated Moir\'e images, and
use it as an auxiliary image set to create a PixMix augmented
dataset; then we train a deep learning model on this
augmented dataset and measure the robustness of its image classification.

Our proposal of MoireDB offers several key advantages, including the following.
\begin{itemize}
      \item The use of MoireDB for data augmentation
            improves robustness with respect to real-world degradations.
      \item MoireDB contains only formula-generated images,
            eliminating copyright problems and making the database
            suitable for commercial use.
      \item The images constituting MoireDB are auto-generated, reducing
            the cost of assembling images into datasets.
\end{itemize}

%% file: figs/en-fig1.tex

\begin{figure}[t]
    \centering
    \includegraphics[width=8.5cm]{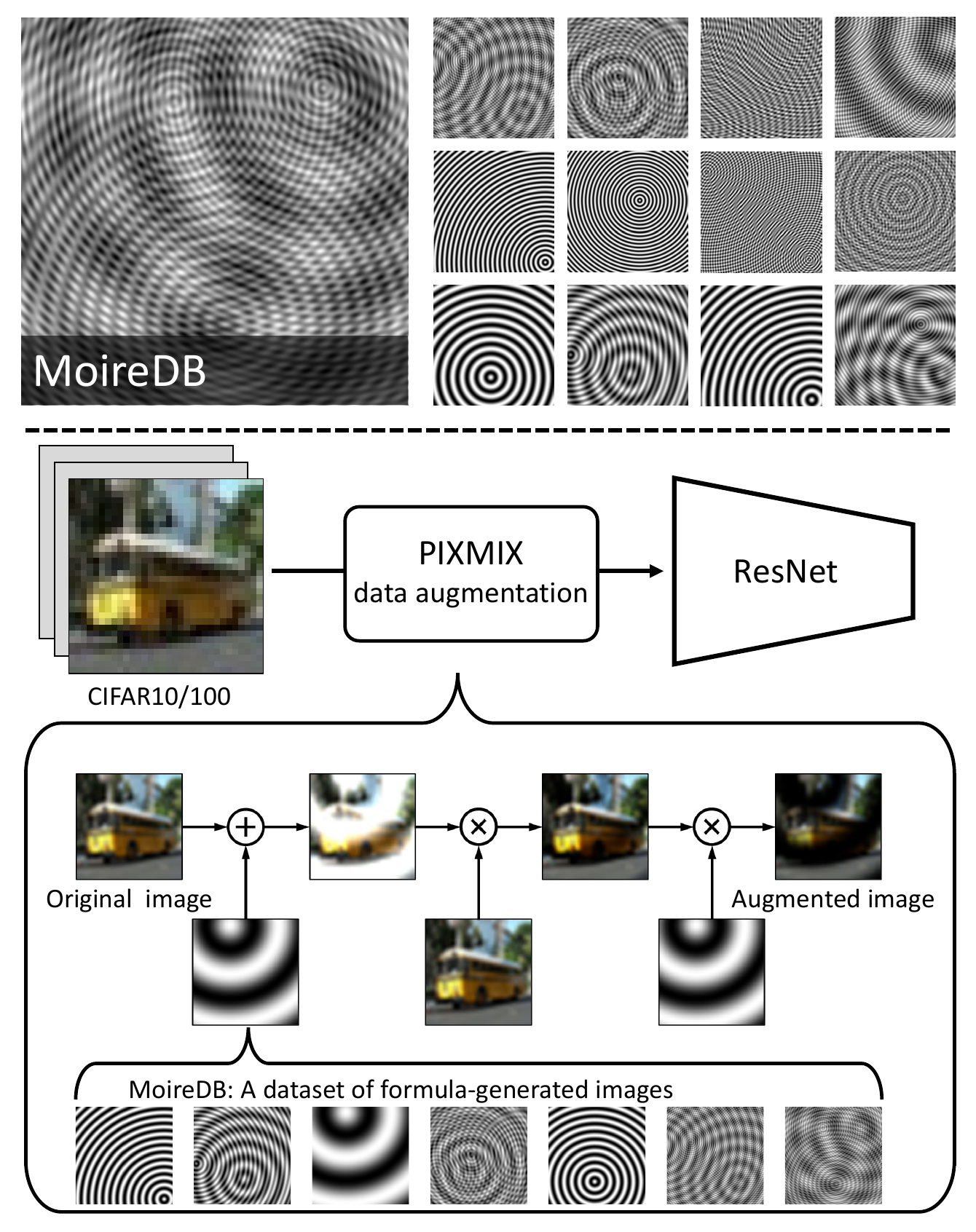}
    \vspace{-12pt}
    \caption{Data augmentation methods based on formula-generated images~\cite{ImageNet-C,PIXMIX}.}
    \label{fig:figPIXMIXmoireConcept}
    \vspace{-12pt}
\end{figure}

%% file: 02_related.tex
\section{Related Work}
\label{sec:related}

\subsection{Robustness in image classification}
\label{sec:Robustness}

Digital images are susceptible to noise,
compression, and other sources of corruption
caused by a broad range of mechanisms.
Although such corruption does not prevent
human visual perception from identifying
images with high accuracy, it \textit{does}
significantly reduce the image identification accuracy
of image recognition models~\cite{ImageNet-C},
and improving the robustness of image recognition models
is a central challenge for image recognition research.

The robustness of image recognition models
may be quantified by testing on specialized
datasets such as ImageNet-C and CIFAR-C, which
consist of images that have been corrupted in various
ways---such as by adding noise, blurring, weathering,
or applying digital transformations---to reflect 15
types of corruption commonly experienced by
digital images; as an example, one corrupted image 
from CIFAR-C is shown in Fig.~\ref{figCIFARC}.

To quantify robustness using ImageNet-C or CIFAR-C,
the image classification accuracy is measured for each of the
15 categories of image corruption, and an average is performed over all
categories to yield a \textit{mean corruption error} (mCE);
smaller mCE values indicate greater robustness.

In addition to quantifying robustness against corruption,
robustness may also be quantified against adversaries, 
i.e., adversarial attacks, by measuring
image classification accuracy for special test images
in the ImageNet and CIFAR datasets to which adversarial
attacks have been applied; again, lower values of the 
image classification accuracy indicate greater robustness.

\subsection{Robustness and illusory images}
\label{sec:Illusion_Moire_Robust}
Image recognition models have been often shown to be
affected by illusory images in the same ways humans are. For example,
illusions that confuse viewers into making erroneous color 
judgments~\cite{color2}, and illusions that trick viewers into 
perceiving a boundary contour that is not actually present~\cite{abu} 
provoke responses from deep learning models that resemble the responses 
of human viewers~\cite{color1,illusion}.
Illusory studies investigating the behavior of image recognition models
on such illusory images often include descriptions of tests
to characterize robustness; for example, studies of color-related 
illusions use corrupted images---obtained by subjecting test images 
to attacks mimicking the image representations responsible for
inducing illusions---to test the behavior of deep learning 
models~\cite{color1}. The results of such tests indicate improved
robustness for CNNs capable of accounting for illusions.
Similarly, studies of boundary-contour illusions have
discussed tests to assess the robustness of deep learning models
capable of taking illusions into consideration~\cite{abu}.

\input{figs/en-fig2}

One common type of illusion involves static images
that viewers erroneously perceive to be in motion;
illusions of this type often feature concentric circles 
or striped patterns~\cite{abu,Fraser}. One
well-known class of images of this type is composed of images
incorporating interference fringes; as mentioned above, we refer to such images as Moir\'e images.
Moir\'e images are produced by superposing
simple patterns of stripes, concentric circles,
or similar elements. They are closely related to
optical illusions and have been studied in fields
ranging from image processing to visual art~\cite{moire_illusion}.

\input{figs/en-fig3}

Moir\'e patterns arise naturally in digital images;
for example, interference between a striped texture pattern
and the spatial frequencies present in a digital image
can produce Moir\'e textures.
Within the field of image recognition, researchers have studied
techniques for eliminating Moir\'e features in digital images
and for creating image recognition models capable of recognizing
the emergence of Moir\'e features~\cite{Moire, MoireCNN}.
In the latter case, the creation of Moir\'e-aware image recognition
models has been shown to improve robustness against
image corruption~\cite{MoireCNN}.

These observations suggest that illusory images and Moir\'e images
may have significant ramifications for robustness---and motivate the
basic assumption on which the present study is premised. Namely, we hypothesize
that using Moir\'e images for data augmentation will improve the 
robustness of deep learning models trained on the resulting
augmented datasets.

\subsection{PIXMIX}
\label{sec:PIXMIX}

In the PixMix approach to data augmentation,
training images from databases such as ImageNet or CIFAR
are combined additively or multiplicatively with an auxiliary set
of structurally complex images to yield an augmented dataset;
deep learning models trained on the augmented dataset then
exhibit improved image identification accuracy and robustness
compared to models trained on the non-augmented dataset.
In the original PixMix proposal, the auxiliary set of 
structurally complex images included two types of images:
Fractal arts and FVis. Examples of these two
types of images are shown in Fig~\ref{figMixImages}.

Fractal arts (note that this is different from FractalDB) are manually designed images downloaded
from DeviantArt; these images contain
shapes and color schemes designed to pique the curiosity
of human visual perception, and are thus expected to be
structurally complex.
FVis are machine-generated images that may be downloaded
from OpenAI Microscope. This database allows
visualization results for image features---as extracted
by various pre-trained CNN models operating on a large image
dataset---to be downloaded in the form of image files.
The structural complexity of these feature-visualization images
is often comparable to that of Fractal arts.

Given an input image dataset, PixMix produces an
augmented dataset by performing repeated mixing operations.
Specifically, each input image is subjected to
a randomly chosen number (at most 5) of mixing steps and 
in each step, the image is mixed either with an input
image or with an image chosen from the auxiliary image
set, and the mixing is performed either additively or
multiplicatively (chosen at random).
Deep learning models trained on PixMix augmented datasets
are known to exhibit improved image identification accuracy 
and robustness compared to other data augmentation methods
such as Mixup~\cite{Mixup} or CutMix~\cite{CutMix}.

However, some Fractal arts are protected by
copyright, and thus commercial use of PixMix
remains questionable.
Moreover, both Fractal arts and FVis are enormously
costly to generate, and the number of images
that may be feasibly assembled into a dataset
is limited in practice.

All of these problems may be eliminated by using
formula-generated image datasets. Examples include
DeadLeaves and FractalDB, discussed in detail in Section 2.3.
However, data augmentation using the formula-generated images
of FractalDB~\cite{FractalDB} and DeadLeaves (Squares)~\cite{DeadLeaves}
is known to be less effective than data augmentation
using Fractal arts and FVis.

Therefore, in the present study, we propose, investigate, and
evaluate the performance of a new strategy for
data augmentation using formula-generated images
that promises image identification accuracy and robustness
comparable to or greater than that of data augmentation
using Fractal arts and FVis.

\input{figs/en-fig4}

\subsection{Formula-driven Supervised Learning (FDSL)}
\label{sec:FDSL}

In formula-driven supervised learning (FDSL),
a large image dataset consisting of formula-generated images
is used to pre-train a image recognition model.

Pre-training of image recognition models on large-scale image datasets
is known to yield significant improvements in image identification accuracy
for additional training~\cite{DeCAF,wmi,dbi}.
The pre-training with large-scale image datasets is typically chosen to be ImageNet, which contains over 14 million real-world images.

However, ImageNet and other large-scale image datasets collected on the Internet
cannot be used commercially, because the images they
contain are subject to copyright protections and
privacy concerns~\cite{ImageNet,Gender,Fairer}.
The large amounts of time and manpower required to
generate annotations by hand, as well as the high cost of
assembling images into datasets, also render this approach impractical.
FDSL eliminates problems of usage rights
and of costly dataset construction~\cite{FractalDB};
because the formula-generated image datasets constructed
and used in FDSL methods consist of copyright-free
images, they---unlike ImageNet---may be used to train deep learning
models intended for commercial use.

One proposed strategy for constructing
formula-generated image datasets is that
of the FractalDB, and image recognition
models pre-trained on FractalDB are known to yield improved
image recognition accuracy during additional training,
as is true for models pre-trained on ImageNet.
The formulas used to generate images in FractalDB
are based on fractal geometry, and
a wide variety of shapes and patterns may be drawn
by varying the adjustable parameters in these formulas.

As of 2024, the FDSL framework has been extended for representations (e.g., tiling~\cite{TileDB}, contours~\cite{RCDB}, Perlin noise~\cite{PerlinNoiseDB}), modalities (e.g., video~\cite{VideoPerlin}, multi-view~\cite{MV-FractalDB}, point cloud~\cite{PC-FractalDB}), and tasks (e.g., segmentation~\cite{SegRCDB}, limited pre-training~\cite{OFDB,1p-frac}, 3D segmentation~\cite{PrimGeoSeg}). 

Here, the current FDSL dataset boasting the
greatest pre-training efficacy is VisualAtom~\cite{VisualAtom}.
Images in VisualAtom, like images in FractalDB,
can be made to incorporate a wide variety
of shapes and patterns by varying adjustable parameters
in the image-generation formulas.
The formulas used to generate VisualAtom images
can also produce images featuring outlines of
complex and diverse shapes.
In PixMix-based approaches,
auxiliary images of greater structural complexity
are known to be more effective for data augmentation.
This explains why the formula-generated images
constituting FractalDB and VisualAtom have proven themselves useful in practice.

For these reasons, in discussing the results of tests to
evaluate the performance of the novel technique proposed
herein (Section~\ref{sec:evaluate}),  we will
contextualize this performance by comparing it
to the observed performance of data augmentation
using FractalDB (as reported in the original PixMix paper)
and to that of data augmentation using VisualAtom.

%% file: figs/en-fig2.tex

\begin{figure}
    \centering
    \includegraphics[width=8.0cm]{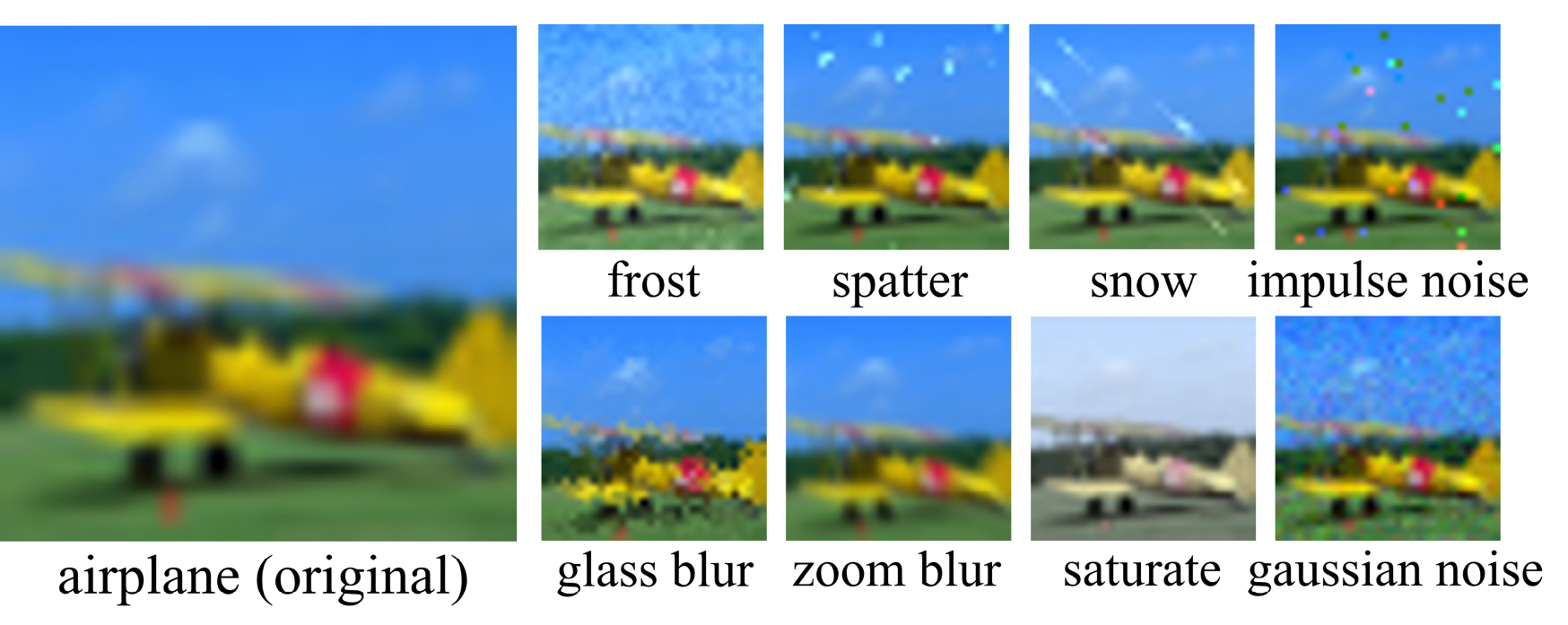}
    \vspace{-12pt}
    \caption{CIFAR-C~\cite{ImageNet-C}.}
    \label{figCIFARC}
\end{figure}

%% file: figs/en-fig3.tex

\begin{figure}
    \centering
    \includegraphics[width=8.0cm]{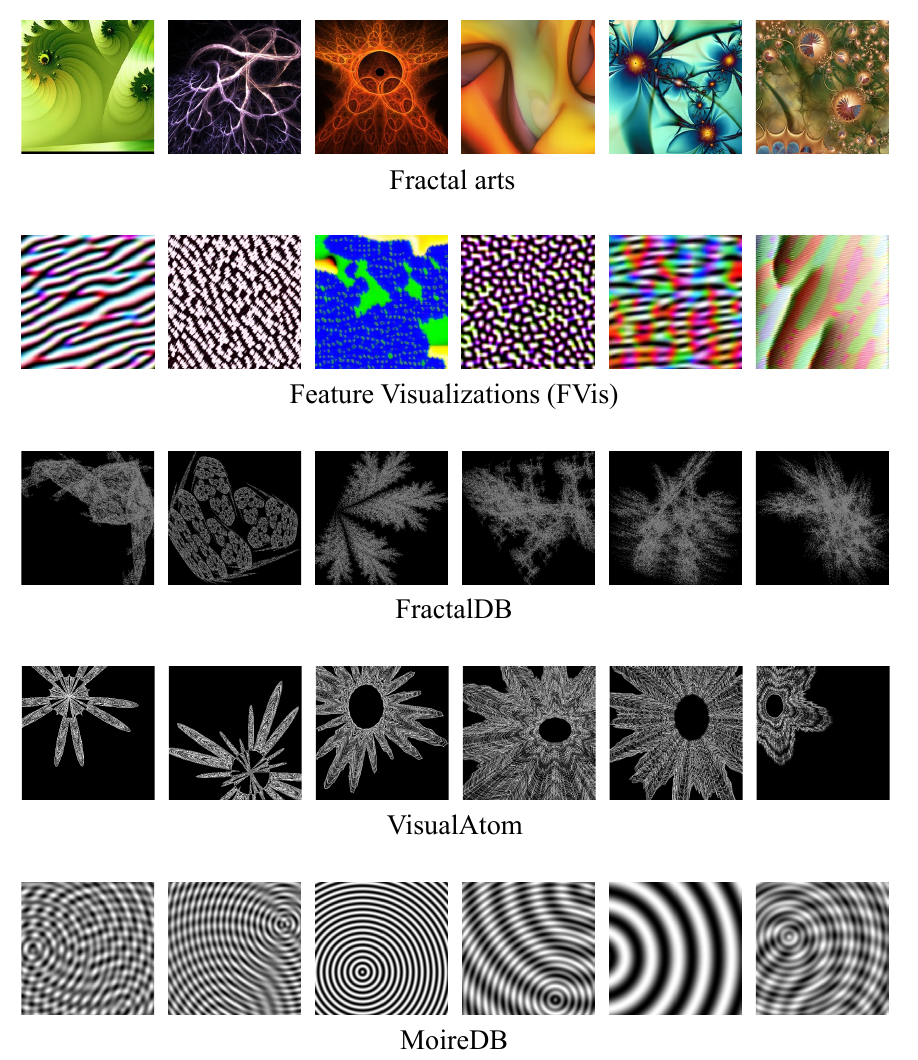}
    \caption{Examples of data augmentation images~\cite{PIXMIX,FractalDB,VisualAtom}.}
    \label{figMixImages}
\end{figure}

%% file: figs/en-fig4.tex

\begin{figure*}
    \centering
    \includegraphics[width=16.0cm]{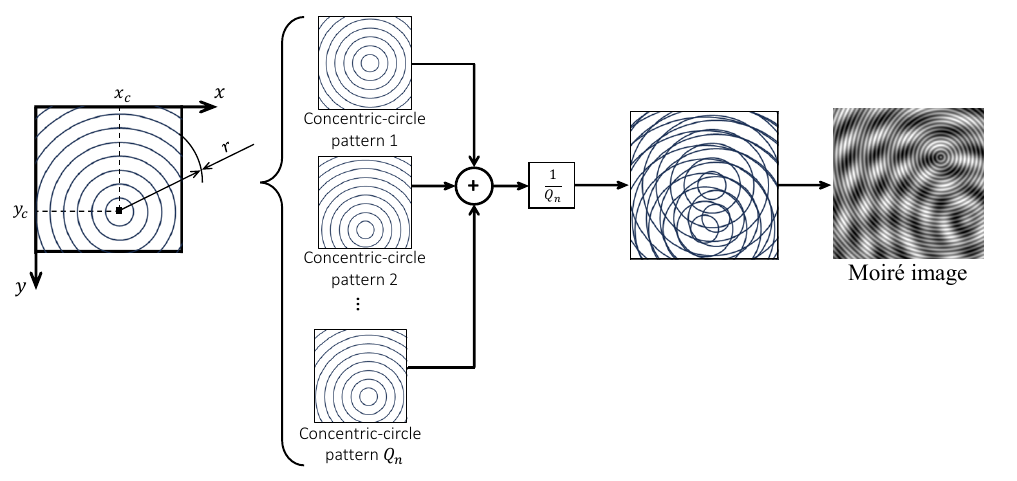}
    \vspace{-12pt}
    \caption{Our algorithm for generating Moir\'e images.}
    \label{figMoire}
\end{figure*}

%% file: 03_method.tex
\section{Data augmentation using formulagenerated Moir´e images}
\label{sec:method}

In the present study, we propose MoireDB, a formula-generated image
dataset for data augmentation. Our goal is to ensure that
training on MoireDB-augmented image datasets
increases the robustness of image recognition models
for classification tasks. Section~\ref{sec:MoireDB-Render}
describes our procedure for generating the Moir\'e images
comprising MoireDB, while Section~\ref{sec:MoireMIX-Render}
discusses our strategy for data augmentation using
generated Moir\'e images.

As noted above, the idea of generating Moir\'e images
for use in data augmentation is motivated by the hypothesis
that using and against illusory images for data augmentation should
improve robustness.

\subsection{Generation of Moir\'e images}
\label{sec:MoireDB-Render}

\input{figs/en-fig5}

\begin{table*}[t]
    \vspace{12pt}
    \caption{Adjustable parameters for auto-generated Moir\'e images}
    \centering
    \vspace{-8pt}
    \begin{tabular}{lc|c} \toprule[0.8pt]
        Parameter                    & Symbol     & Range \\\midrule[0.8pt]
        Interval frequency           & $\nu$      & $0.01\leq \nu < 0.05$ \\[1ex]
        Center-point coordinates           & $x_c, y_c$ & $0\leq (x_c, y_c) < 600$  \\[1ex]
%
         Number of superposed concentric-circle patterns & $Q_n$       & $Q_n = \{1, 2, 3\}$  \\
        \bottomrule[0.8pt] \\
    \end{tabular}
    \vspace{-20pt}
    \label{tab:MoireDBparam}
\end{table*}

Our algorithm for generating the Moir\'e images constituting
MoireDB is depicted schematically in Fig.~\ref{figMoire}.
The starting point is a simple procedure
(Fig.~\ref{figMoire}, far left) for generating a
concentric-circle pattern; this procedure is described
by a formula, discussed below, containing multiple
adjustable parameters such as the coordinates 
$(x_c,y_c)$ of the common center point.
To generate a single Moir\'e image, we invoke this 
formula multiple times---with randomly chosen values
for the adjustable parameters---to yield a set 
of multiple distinct concentric-circle patterns 
(Fig.~\ref{figMoire}, center), then simply \textit{superpose}
these to yield the Moir\'e image (Fig.~\ref{figMoire}, far right).
The superposition of randomly generated
concentric-circle patterns gives rise to the characteristic
interference fringes of Moir\'e images, and varying the
adjustable parameters defining the concentric-circle
patterns allows a wide range of distinct fringe patterns to be
realized.

The image representing each concentric-circle pattern is generated
by a formula that computes a brightness value for each pixel in the image.
Each Moir\'e image depends on several adjustable parameters:
the number $Q_n$ of concentric-circle patterns superposed, and,
for each of these patterns, the center-point coordinates $(x_c, y_c)$
and an interval frequency parameter $\nu$ described below.
Values for all of these parameters are chosen randomly within
the ranges listed in Table~\ref{tab:MoireDBparam}.
 
Each concentric-circle pattern may be described as a superposition
of circles of the form

\begin{equation}
f_{Q_n}=\frac{1}{Q_n}\sum^{m}_{k=1}\eta_k\in\mathbb{R}^2
\end{equation}
where $m$ is the number of circles drawn in the pattern
and $\eta_k$ represents the $k$-th circle. Denoting the radius of 
this circle by $r_k$, and recalling that the circle is centered
at $(x_c,y_c)$, we may express $\eta_k$ in the form 

\begin{equation}
  \eta_k=
  \left\{
    \begin{array}{l}
      x=(r_k\cos{\theta}+x_c)\times g \\
      y=(r_k\sin{\theta}+y_c)\times g
    \end{array}
  \right.
  (0\leq\theta<2\pi)
\end{equation}
The center-point coordinates ($x_c$,$y_c$) are chosen at random from 
a uniform distribution.
The quantity $g$ in this expression, representing the brightness
at point $(x,y)$, is a sinusoidally varying function of the
radial distance $r$:

\begin{equation}
g=(V_M(cos(\nu \times \pi \times r)) + 1)\times255,
\end{equation}
where $V_M$ is the amplitude of the sinusoidal brightness variation.
Using the brightness $g$ to define a grayscale value for each pixel
yields an image representing the concentric-circle pattern. Choosing
the number of concentric-circle patterns $Q_n>1$ then ensures
interference between the patterns, yielding the desired Moir\'e image.

We set the size of generated images to be $512\times 512$ [pixel];
the number of circles $m$ drawn for each concentric-circle pattern
is determined as appropriate based on the image size and the 
interval frequency $\nu$.

\tabA
\tabNoise
\tabBlur

\subsection{Data augmentation using Moir\'e images}
\label{sec:MoireMIX-Render}

Our strategy for data augmentation using Moir\'e images
is outlined schematically in Fig.~\ref{figPIXMIXmoireConcept},
and Fig.~\ref{figPIXMIXmoire} shows a detailed diagram
of the operational pipeline of our PixMix implementation
with Moir\'e images, in this case for an example involving
1 additive mixing operation and 2 multiplicative mixing
operations. Our data augmentation procedure is the same as
that used in PixMix.
The number of Moir\'e images we generate for data augmentation
is 14,230, chosen to match the number of Fractal arts used in PixMix.
For each image, the parameter values in the image-generation
formulas are chosen at random from the ranges listed in
Table~\ref{tab:MoireDBparam}.
For each mixing step, we choose an image at random from the 
set of generated images and mix it either additively or 
multiplicatively with the selected Moir\'e image or with 
the input image.

\tabWea
\tabDig

%% file: figs/en-fig5.tex

\begin{figure*}[t]
    \vspace{-12pt}
    \centering
    \includegraphics[width=14.0cm]{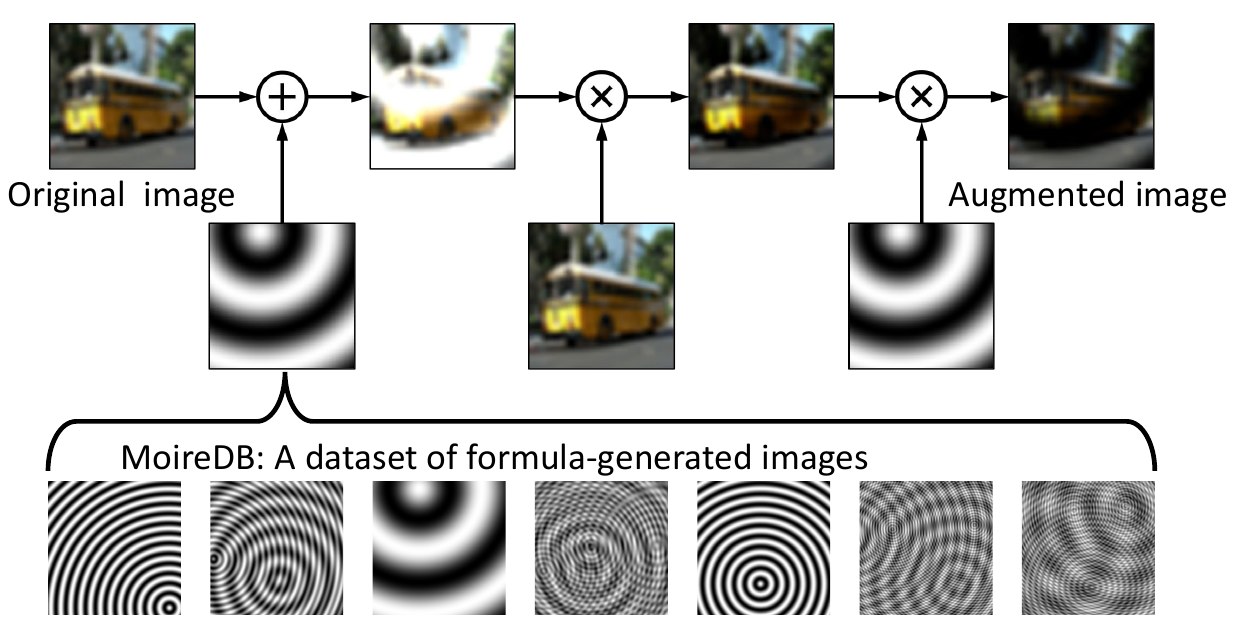}
    \vspace{-12pt} 
    \caption{Example illustrating PIXMIX-style data augmentation
             with Moir\'e images~\cite{ImageNet-C,PIXMIX}.}
    \label{figPIXMIXmoire}
    \vspace{-12pt}
\end{figure*}

%% file: 04_experimental.tex
\section{Experimental evaluation}
\label{sec:experimental}

\subsection{Test procedure}
\label{sec:experiment}
We conducted experimental tests to assess the effectiveness
of data augmentation using MoireDB, comparing the results
against robustness values obtained via several alternative
models: data augmentation using Fractal arts and FVis, as originally
proposed for PixMix, and PixMix with data augmentation
images taken from FractalDB and VisualAtom.

The training model we use is WideResNet~\cite{PIXMIX,WideResNet}.
We use CIFAR as a training-image dataset.
For each of the various data augmentation strategies, we create
an augmented version of the CIFAR training-image dataset,
then train WideResNet on the augmented dataset for 100 epochs
and measure the robustness of the trained model.
Robustness is measured on the CIFAR-C dataset of test images
using the Corruptions and Adversaries evaluation tasks~\cite{PIXMIX}.

The Corruptions task involves using CIFAR-C to measure
robustness against image corruption~\cite{ImageNet-C}.
The metric for this assessment is the previously mentioned mCE, which is smaller for greater robustness.
mCE is computed as the mean image identification accuracy
for the 15 types of image corruption represented by CIFAR-C.

The Adversaries task involves measuring robustness
against adversarial attack~\cite{Adrobust}.
The metric for this assessment is the image identification accuracy,
with lower values indicating better performance.
Adversarial attacks are applied to CIFAR test images.

\subsection{Results of robustness tests}
\label{sec:evaluate}

Table~\ref{tabA} shows the results of tests to assess
the impact of MoireDB-based data augmentation on
the robustness of image classification.
The column labeled ``Baseline'' lists
results from the original PixMix paper~\cite{PIXMIX}.

From Table~\ref{tabA} we see that 
data augmentation using MoireDB achieves better
image identification robustness than any other method---including
data augmentation using Fractal arts---for both CIFAR-10-C and 
CIFAR-100-C.
Comparing results for the FDSL datasets 
FractalDB, VisualAtom, and MoireDB,
we see that, in \textit{every} test of robustness,
the largest robustness improvement
is achieved for data augmentation using MoireDB.

These results demonstrate that MoireDB-based data augmentation
can yield robustness improvements comparable to or greater than
data augmentation using Fractal arts or FVis.

To analyze the test results in greater detail,
we consider image classification accuracies
for the various types of image corruption in CIFAR-100-C.
From Table~\ref{tabNoise} we see that, for
all forms of image corruption caused by noise,
the greatest improvement in image classification robustness
is achieved by data augmentation using FVis.
On the other hand, from Table~\ref{tabBlur} we see that,
for various forms of blurring,
MoireDB tends to yield greater robustness improvements
than other image datasets. According to Table~\ref{tabBlur}, we can see MoireDB with PixMix performaed better results on the blurred noise types on the validation of CIFAR-100-C dataset.

Similarly, from Tables~\ref{tabWea} and~\ref{tabDig}
we see that, for image corruption due to snow or frost,
as well as for image corruption due to elastic deformation
or pixelation, data augmentation using MoireDB achieves
the greatest improvement in robustness.

%% file: 05_conclusions.tex
\section{Conclusion}
\label{sec:conclusion}

In the present study, we proposed MoireDB,
a formula-generated dataset of interference-fringe images
for use with the PixMix method of data augmentation,
and conducted experiments to assess its impact on robustness.
Our results showed that, for several test categories,
data augmentation using MoireDB
achieved a greater improvement in robustness than
data augmentation with Fractal arts or FVis.
This demonstrates that formula-generated images based on illusory images
can help improve the robustness of deep learning models for image classification.

%% file: 12_appendix.tex
\section{Appendix Section}
\label{sec:appendix_section}
Supplementary material goes here.

%% file: _main.bbl
\begin{thebibliography}{39}
\providecommand{\natexlab}[1]{#1}
\providecommand{\url}[1]{\texttt{#1}}
\expandafter\ifx\csname urlstyle\endcsname\relax
  \providecommand{\doi}[1]{doi: #1}\else
  \providecommand{\doi}{doi: \begingroup \urlstyle{rm}\Url}\fi

\bibitem[Baradad et~al.(2021)Baradad, Wulff, Wang, Isola, and Torralba]{DeadLeaves}
Manel Baradad, Jonas Wulff, Tongzhou Wang, Phillip Isola, and Antonio Torralba.
\newblock Learning to see by looking at noise.
\newblock \emph{arXiv preprint arXiv:2106.05963}, 2021.

\bibitem[Buolamwini and Gebru(2018)]{Gender}
Joy Buolamwini and Timnit Gebru.
\newblock Gender shades: Intersectional accuracy disparities in commercial gender classification.
\newblock \emph{ACM FAccT}, 2018.

\bibitem[Deng et~al.(2009)Deng, Dong, Socher, Li, Li, and Fei-Fei]{ImageNet}
Jia Deng, Wei Dong, Richard Socher, Li-Jia Li, Kai Li, and Li Fei-Fei.
\newblock Imagenet: A large-scale hierarchical image database.
\newblock In \emph{CVPR}, 2009.

\bibitem[Donahue et~al.(2014)Donahue, Jia, Vinyals, Hoffman, Zhang, Tzeng, and Darrell]{DeCAF}
Jeff Donahue, Yangqing Jia, Oriol Vinyals, Judy Hoffman, Ning Zhang, Eric Tzeng, and Trevor Darrell.
\newblock Decaf: A deep convolutional activation feature for generic visual recognition.
\newblock In \emph{ICML}, 2014.

\bibitem[Fan and Zeng(2022)]{abu}
Jinyu Fan and Yi Zeng.
\newblock Abutting grating illusion: Cognitive challenge to neural network models.
\newblock \emph{arXiv preprint arXiv:2208.03958}, 2022.

\bibitem[Gomez-Villa et~al.(2019{\natexlab{a}})Gomez-Villa, Fernández, Vazquez-Corral, Bertalmío, and Malo]{color1}
Alex Gomez-Villa, Adrián~Martín Fernández, Javier Vazquez-Corral, Marcelo Bertalmío, and Jesús Malo.
\newblock Visual illusions also deceive convolutional neural networks: Analysis and implications.
\newblock \emph{arXiv preprint arXiv:1912.01643}, 2019{\natexlab{a}}.

\bibitem[Gomez-Villa et~al.(2019{\natexlab{b}})Gomez-Villa, Martin, Vazquez-Corral, and Bertalmio]{color2}
Alexander Gomez-Villa, Adrian Martin, Javier Vazquez-Corral, and Marcelo Bertalmio.
\newblock Convolutional neural networks can be deceived by visual illusions.
\newblock In \emph{CVPR}, 2019{\natexlab{b}}.

\bibitem[He et~al.(2019)He, Wang, Shi, and Duan]{Moire}
Bin He, Ce Wang, Boxin Shi, and Ling-Yu Duan.
\newblock Mop moire patterns using mopnet.
\newblock In \emph{ICCV}, 2019.

\bibitem[He et~al.(2016)He, Zhang, Ren, and Sun]{resnet}
Kaiming He, Xiangyu Zhang, Shaoqing Ren, and Jian Sun.
\newblock Deep residual learning for image recognition.
\newblock In \emph{CVPR}, 2016.

\bibitem[Hendrycks and Dietterich(2019)]{ImageNet-C}
Dan Hendrycks and Thomas Dietterich.
\newblock Benchmarking neural network robustness to common corruptions and perturbations.
\newblock In \emph{ICLR}, 2019.

\bibitem[Hendrycks et~al.(2022)Hendrycks, Zou, Mazeika, Tang, Li, Song, and Steinhardt]{PIXMIX}
Dan Hendrycks, Andy Zou, Mantas Mazeika, Leonard Tang, Bo Li, Dawn Song, and Jacob Steinhardt.
\newblock Pixmix: Dreamlike pictures comprehensively improve safety measures.
\newblock In \emph{CVPR}, 2022.

\bibitem[Huh et~al.(2019)Huh, Agrawal, and Efros]{wmi}
Minyoung Huh, Pulkit Agrawal, and Alexei~A. Efros.
\newblock What makes imagenet good for transfer learning?
\newblock In \emph{CVPR}, 2019.

\bibitem[Inoue et~al.(2020)Inoue, Yamagata, and Kataoka]{PerlinNoiseDB}
Nakamasa Inoue, Eisuke Yamagata, and Hirokatsu Kataoka.
\newblock Initialization using perlin noise for training networks with a limited amount of data.
\newblock In \emph{ICPR}, 2020.

\bibitem[James(1908)]{Fraser}
Fraser James.
\newblock A new visual illusion of direction.
\newblock \emph{British Journal of Psychology}, 2\penalty0 (3):\penalty0 307--320, 1908.

\bibitem[Kataoka et~al.(2020)Kataoka, Okayasu, Matsumoto, Yamagata, Yamada, Inoue, Nakamura, and Satoh]{FractalDB}
Hirokatsu Kataoka, Kazushige Okayasu, Asato Matsumoto, Eisuke Yamagata, Ryosuke Yamada, Nakamasa Inoue, Akio Nakamura, and Yutaka Satoh.
\newblock Pre-training without natural images.
\newblock In \emph{ACCV}, 2020.

\bibitem[Kataoka et~al.(2021)Kataoka, Matsumoto, Yamagata, Yamada, Inoue, and Satoh]{TileDB}
Hirokatsu Kataoka, Asato Matsumoto, Eisuke Yamagata, Ryosuke Yamada, Nakamasa Inoue, and Yutaka Satoh.
\newblock Formula-driven supervised learning with recursive tiling patterns.
\newblock In \emph{ICCVW}, 2021.

\bibitem[Kataoka et~al.(2022{\natexlab{a}})Kataoka, Hayamizu, Yamada, Nakashima, Takashima, Zhang, Martinez-Noriega, Inoue, and Yokota]{RCDB}
Hirokatsu Kataoka, Ryo Hayamizu, Ryosuke Yamada, Kodai Nakashima, Sora Takashima, Xinyu Zhang, Edgar~Josafat Martinez-Noriega, Nakamasa Inoue, and Rio Yokota.
\newblock Replacing labeled real-image datasets with auto-generated contours.
\newblock In \emph{CVPR}, 2022{\natexlab{a}}.

\bibitem[Kataoka et~al.(2022{\natexlab{b}})Kataoka, Yamagata, Hara, Hayashi, and Inoue]{VideoPerlin}
Hirokatsu Kataoka, Eisuke Yamagata, Kensho Hara, Ryusuke Hayashi, and Nakamasa Inoue.
\newblock Spatiotemporal initialization for 3d cnns with generated motion patterns.
\newblock In \emph{WACV}, 2022{\natexlab{b}}.

\bibitem[Kornblith et~al.(2017)Kornblith, Shlens, and Le]{dbi}
Simon Kornblith, Jonathon Shlens, and Quoc~V. Le.
\newblock Do better imagenet models transfer better?
\newblock In \emph{ICCV}, 2017.

\bibitem[Krizhevsky et~al.(2012)Krizhevsky, Sutskever, and Hinton]{alexnet}
Alex Krizhevsky, Ilya Sutskever, and Geoffrey~E. Hinton.
\newblock Imagenet classification with deep convolutional neural networks.
\newblock In \emph{NeurIPS}, 2012.

\bibitem[Lopes et~al.(2019)Lopes, Poole, Gilmer, and Cubuk]{ImRobust}
Raphael~Gontijo Lopes, Dong~YinandBen Poole, Justin Gilmer, and Ekin~Dogus Cubuk.
\newblock Improving robustness without sacrificing accuracy with patch gaussian augmentation.
\newblock \emph{arXiv preprint arXiv:1906.02611}, 2019.

\bibitem[Madry et~al.(2018)Madry, Makelov, Schmidt, Tsipras, and Vladu]{Adrobust}
Aleksander Madry, Aleksandar Makelov, Ludwig Schmidt, Dimitris Tsipras, and Adrian Vladu.
\newblock Towards deep learning models resistant to adversarial attacks.
\newblock In \emph{ICLR}, 2018.

\bibitem[Nakamura et~al.(2023)Nakamura, Kataoka, Takashima, Noriega, Yokota, and Inoue]{OFDB}
Ryo Nakamura, Hirokatsu Kataoka, Sora Takashima, Edgar Josafat~Martinez Noriega, Rio Yokota, and Nakamasa Inoue.
\newblock Pre-training vision transformers with very limited synthesized images.
\newblock In \emph{ICCV}, 2023.

\bibitem[Nakamura et~al.(2024)Nakamura, Tadokoro, Yamada, Asano, Laina, Rupprecht, Inoue, Yokota, and Kataoka]{1p-frac}
Ryo Nakamura, Ryu Tadokoro, Ryosuke Yamada, Yuki~M. Asano, Iro Laina, Christian Rupprecht, Nakamasa Inoue, Rio Yokota, and Hirokatsu Kataoka.
\newblock Scaling backwards: Minimal synthetic pre-training?
\newblock In \emph{ECCV}, 2024.

\bibitem[Olah et~al.(2017)Olah, Mordvintsev, and Schubert]{distill}
Chris Olah, Alexander Mordvintsev, and Ludwig Schubert.
\newblock Feature visualization, 2017.
\newblock \textit{Distill}, https://distill.pub/2017/feature-visualization.

\bibitem[Shinoda et~al.(2023)Shinoda, Hayamizu, Nakashima, Inoue, Yokota, and Kataoka]{SegRCDB}
Risa Shinoda, Ryo Hayamizu, Kodai Nakashima, Nakamasa Inoue, Rio Yokota, and Hirokatsu Kataoka.
\newblock Segrcdb: Semantic segmentation via formula-driven supervised learning.
\newblock In \emph{ICCV}, 2023.

\bibitem[Simonyan and Zisserman(2014)]{vggnet}
Karen Simonyan and Andrew Zisserman.
\newblock Very deep convolutional networks for large-scale image recognition.
\newblock In \emph{NeurIPS}, 2014.

\bibitem[Spillmann(1993)]{moire_illusion}
Lothar Spillmann.
\newblock The perception of movement and depth in moiré patterns.
\newblock \emph{Perception}, 22:\penalty0 287--308, 1993.

\bibitem[Tadokoro et~al.(2023)Tadokoro, Yamada, Nakashima, Nakamura, and Kataoka]{PrimGeoSeg}
Ryu Tadokoro, Ryosuke Yamada, Kodai Nakashima, Ryo Nakamura, and Hirokatsu Kataoka.
\newblock Primitive geometry segment pre-training for 3d medical image segmentation.
\newblock In \emph{34th British Machine Vision Conference 2023, {BMVC} 2023, Aberdeen, UK, November 20-24, 2023}. BMVA, 2023.

\bibitem[Takashima et~al.(2023)Takashima, Hayamizu, Inoue, Kataoka, and Yokota]{VisualAtom}
Sora Takashima, Ryo Hayamizu, Nakamasa Inoue, Hirokatsu Kataoka, and Rio Yokota.
\newblock Visual atoms: Pre-training vision transformers with sinusoidal waves.
\newblock In \emph{CVPR}, 2023.

\bibitem[Vasconcelos et~al.(2020)Vasconcelos, Larochelle, Dumoulin, Roux, and Goroshin]{MoireCNN}
Cristina Vasconcelos, Hugo Larochelle, Vincent Dumoulin, Nicolas~Le Roux, and Ross Goroshin.
\newblock An effective anti-aliasing approach for residual networks.
\newblock \emph{arXiv preprint arXiv:2011.10675}, 2020.

\bibitem[Watanabe et~al.(2018)Watanabe, Kitaoka, Sakamoto, Yasugi, and Tanaka]{illusion}
Eiji Watanabe, Akiyoshi Kitaoka, Kiwako Sakamoto, Masaki Yasugi, and Kenta Tanaka.
\newblock Illusory motion reproduced by deep neural networks trained for prediction.
\newblock \emph{Frontiers in Psychology}, 2018.

\bibitem[Yamada et~al.(2021)Yamada, Takahashi, Suzuki, Nakamura, Yoshiyasu, Sagawa, and Kataoka]{MV-FractalDB}
Ryosuke Yamada, Ryo Takahashi, Ryota Suzuki, Akio Nakamura, Yusuke Yoshiyasu, Ryusuke Sagawa, and Hirokatsu Kataoka.
\newblock Mv-fractaldb: Formula-driven supervised learning for multi-view image recognition.
\newblock In \emph{IROS}, 2021.

\bibitem[Yamada et~al.(2022)Yamada, Kataoka, Chiba, Domae, and Ogata]{PC-FractalDB}
Ryosuke Yamada, Hirokatsu Kataoka, Naoya Chiba, Yukiyasu Domae, and Tetsuya Ogata.
\newblock Point cloud pre-training with natural 3d structures.
\newblock In \emph{CVPR}, 2022.

\bibitem[Yang et~al.(2020)Yang, Qinami, Fei-Fei, Deng, and Russakovsky]{Fairer}
Kaiyu Yang, Klint Qinami, Li Fei-Fei, Jia Deng, and Olga Russakovsky.
\newblock Towards fairer datasets: Filtering and balancing the distribution of the people subtree in the imagenet hierarchy.
\newblock \emph{ACM FAccT}, 2020.

\bibitem[Yun et~al.(2019)Yun, Han, Oh, Chun, Choe, and Yoo]{CutMix}
Sangdoo Yun, Dongyoon Han, Seong~Joon Oh, Sanghyuk Chun, Junsuk Choe, and Youngjoon Yoo.
\newblock Cutmix: Regularization strategy to train strong classifiers with localizable features.
\newblock In \emph{ICCV}, 2019.

\bibitem[Zagoruyko and Komodakis(2016)]{WideResNet}
Sergey Zagoruyko and Nikos Komodakis.
\newblock Wide residual networks.
\newblock In \emph{BMVC}, 2016.

\bibitem[Zhang et~al.(2017)Zhang, Cisse, Dauphin, and Lopez-Paz]{Mixup}
Hongyi Zhang, Moustapha Cisse, Yann Dauphin, and David Lopez-Paz.
\newblock Mixup: Beyond empirical risk minimization.
\newblock In \emph{ICLR}, 2017.

\bibitem[Zimmermann et~al.(2021)Zimmermann, Borowski, Geirhos, Bethge, Wallis, and Brendel]{FVis}
Roland~S. Zimmermann, Judy Borowski, Robert Geirhos, Matthias Bethge, Thomas S.~A. Wallis, and Wieland Brendel.
\newblock How well do feature visualizations support causal understanding of cnn activations?
\newblock In \emph{NeurIPS}, 2021.

\end{thebibliography}
